\documentclass[11pt, a4paper, abstract=true, oneside, DIV=12]{scrartcl}

\usepackage[T1]{fontenc}
\usepackage[utf8]{inputenc}
\usepackage[USenglish]{babel}
\usepackage[detect-all,binary-units]{siunitx}
\usepackage{amsfonts}
\usepackage{amsmath}
\usepackage{amssymb}
\usepackage{booktabs}
\usepackage{graphicx}
\usepackage{subfig}
\usepackage{relsize}
\usepackage[numbers]{natbib}
\usepackage{authblk}
\usepackage{hyperref}
\usepackage{tabularx}
\newcolumntype{Y}{>{\raggedleft\arraybackslash}X}
\usepackage[boxed]{algorithm}
\usepackage[noend]{algpseudocode}

\makeatletter
\DeclareRobustCommand\Cpp{C\raisebox{1pt}{{\relsize{-2}++}}\ }
\makeatother

\makeatletter
\renewcommand*{\NAT@spacechar}{~}
\makeatother

\newcommand{\abs}[1]{\ensuremath{\left\vert#1\right\vert}}

\newcommand{\bigO}{\ensuremath{\mathcal{O}}}

\newcommand{\pixelspace}{\ensuremath{\mathcal{P}}}
\newcommand{\neighborhoodsystem}{\ensuremath{\mathcal{N}}}
\newcommand{\imagesymbol}{\ensuremath{I}}
\newcommand{\image}[1]{\ensuremath{\imagesymbol_{#1}}}
\newcommand{\segmentationsymbol}{\ensuremath{A}}
\newcommand{\segmentation}[1]{\ensuremath{\segmentationsymbol_{#1}}}
\newcommand{\dist}{\ensuremath{d}}

\newcommand{\partitioning}[1]{\ensuremath{\mathcal{R}_{#1}}}
\newcommand{\component}[1]{\ensuremath{C_{#1}}}
\newcommand{\score}[1]{\ensuremath{s(C_{#1})}}
\newcommand{\event}[1]{\ensuremath{X_{#1}}}

\newcommand{\prob}[1]{\ensuremath{\mathrm{P}(#1)}}

\DeclareSIUnit[number-unit-product = {}]\pixel{P}

\algnewcommand{\algorithmicreturns}{\textbf{returns}}
\algrenewcommand{\algorithmiccomment}[1]{\hskip3em$//$ #1}
\algrenewcommand{\algorithmicforall}{\textbf{for each}}
\algrenewcommand\Return{\State \algorithmicreturn{} }%
\algdef{SE}[FUNCTION]{Function}{EndFunction}%
   [3]{\algorithmicfunction\ \textproc{#1}\ifthenelse{\equal{#2}{}}{}{(#2)}\ \ifthenelse{\equal{#3}{}}{}{\algorithmicreturns\ #3}}%
   {}%
\algtext*{EndFunction}% works only for noend

\title{3D Cell Nuclei Segmentation with Balanced Graph Partitioning}
\author[1]{Julian Arz}
\author[1]{Peter Sanders}
\author[2]{Johannes Stegmaier}
\author[2]{Ralf Mikut}
\affil[1]{\small Institute for Theoretical Informatics, KIT, Karlsruhe, Germany.}
\affil[2]{\small Institute for Applied Computer Science, KIT, Karlsruhe, Germany.}
\date{\vspace{-5ex}}

\begin{document}

\maketitle
\begin{abstract}
Cell nuclei segmentation is one of the most important tasks in the analysis of biomedical images. With ever-growing sizes and amounts of three-dimensional images to be processed, there is a need for better and faster segmentation methods. Graph-based image segmentation has seen a rise in popularity in recent years, but is seen as very costly with regard to computational demand. We propose a new segmentation algorithm which overcomes these limitations. Our method uses recursive balanced graph partitioning to segment foreground components of a fast and efficient binarization. We construct a model for the cell nuclei to guide the partitioning process. Our algorithm is compared to other state-of-the-art segmentation algorithms in an experimental evaluation on two sets of realistically simulated inputs. Our method is faster, has similar or better quality and an acceptable memory overhead.
\end{abstract}

%------------------------------------------------------------------------- 
\section{Introduction}
\label{sec:intro}
% ------------ % 
% introduction %
% ------------ %
The ultimate objective of developmental biology is to understand the development of a single symmetric cell to a complex three-dimensional organism. A big step towards this goal are recent advances in light microscopy technology, which have paved the way for the large-scale \textit{in vivo} investigation of biological structure.
For an example, the experiments of \citet{tomer12quantitative} on the embryonic development of \textit{Drosophilia} result in more than one million images with a total size of about \SI{10}{\tera\byte} per embryo.

Already the amounts and sizes of these images make it difficult for them to be manually analyzed. Another challenge is posed by the imaging quality. The images suffer from the presence of noise, low contrast and a blurring effect induced by the physical properties of the optical system. When combining the image stacks to three-dimensional images, a degradation of image quality with increasing axial depth can be observed. This phenomenon is due to the scattering of light in living tissue and shadowing of tightly clustered objects. Another obstacle which has to be overcome is the anisotropic resolution. The resolution\footnote{i.e. the number of voxels per unit of length} in axial direction is typically lower than in lateral direction, by a factor of three to five. This renders it difficult to reconstruct a three-dimensional boundary of an object.

To overcome these challenges, automated methods have to be developed which process and analyze these images. The automation also has the advantage that it frees the image analysis process from the researcher's subjective bias and guarantees the reproducibility of the results.
Typically, image processing methods consist of a pipeline of multiple tools: from multiview image restoration to object segmentation, annotation and tracking to the modeling of entire processes. Each stage relies on the results of the previous ones. It is important that each tool have results of high quality, despite the aforementioned image deficiencies. Furthermore, it is natural to demand that the total time to analyze the images is in the order it takes for them to be taken, such that computation time does not evolve into a bottleneck of developmental biology research. Therefore, new algorithms have to be efficient, both in quality and in time.

One of the most important steps in nuclei image analysis is \emph{segmentation}.
The goal of segmentation is to accurately delineate the boundaries of all observed cell nuclei. It translates the pixel-based representation of the data to an object-based one \cite{khairy11reconstructing} and is therefore crucial for downstream tasks such as cell state annotation, cell tracking or lineaging.

Many multi-purpose image segmentation algorithms are based on a graph representation of the image \cite{shi00normalized, boykov06graphcuts}. 
By mapping each pixel to a node in a graph, the image segmentation problem can be reduced to graph theoretic algorithms.
The drawback of the employment of the graph-based techniques is their high memory overhead, which, in the context of cell segmentation, makes them suffer most from the steady increase in the size and the amount of images to be processed. 

In our work, we develop a new graph-based 3D cell nuclei segmentation algorithm which overcomes the current limitations of these methods. We extract foreground regions from the background with a fast and effective binarization. 
We perform graph-based segmentation successively on each connected foreground component. This approach solves the problem of the graph representation induced memory overhead, because the size of these components does not directly depend on the image size.
The splitting phase uses a probabilistic model based on domain knowledge to make decisions.
We perform an experimental evaluation on two sets of realistically simulated inputs. The experiments confirm that our algorithm is both efficient and allows a high quality segmentation even of tightly clustered cells.

\paragraph{Confocal Fluorescence Microscopy.}
Biological structures at cell level are mostly transparent. Therefore, fluorescent markers, such as the green fluorescent protein (GFP), are introduced into the structures to be observed. When illuminated with an excitation wavelength, the fluorophore emits light of a longer wavelength. In conventional fluorescent microscopes, the entire specimen is illuminated. This poses a problem because emitted light from out-of-focus areas reduces the signal contrast from the in-focus plane \cite{inoue06foundations}. 

Confocal laser-scanning microscopy (CLSM) \cite{minsky61microscopy} limits the field of illumination to the region in focus. A point of light scans the specimen by moving over all planes, lines and points. This allows for optical sectioning of an organism, i.e. creating a 3D image.
Confocal fluorescence microscopy was among the most frequently used techniques in the last decades.
However, this technique still suffers from the limited accessible depths, because the laser has to penetrate the entire specimen. Another problem is the effect of photobleaching, a process in which the markers are left unable to fluoresce. These limitations hinder its service for long-time detailed imaging of developing organisms.

In contrast, in selective plane illumination microscopy (SPIM) \cite{huisken04optical}, the entire in-focus plane of the sample is illuminated orthogonally to the detection axis. This technique increases the feasible depth of the sample, as the illumination light does not have to penetrate the sample up to the in-focus plane. It also increases the attainable axial resolution. Furthermore, only the current plane is affected by photobleaching.
A further reduction in phototoxicity is achieved with digital scanned laser light sheet fluorescence microscopy (DSLM) \cite{keller08reconstruction}. These advances make it possible to image organisms over long periods of time, with high spatiotemporal resolutions, enabling new studies of cell behavior in complex developing specimens \cite{khairy11reconstructing}.
For an example, \citet{keller08reconstruction} record entire zebrafish and analyze cell nuclei positions and movement over the first 24 hours of embryogenesis. The study produces stacks of about 400 images with a size of \SI{8}{\mega\pixel} and a temporal resolution of up to \SI{60}{\second}. \citet{tomer12quantitative} combine light-sheet microscopy with multi-view imaging. Their experiments on the embryonic development of \textit{Drosophilia}
result in more than one million images with a total size of about \SI{10}{\tera\byte} per embryo.

% ------------ % 
% related work %
% ------------ %
\subsection{Related Work}
\label{sec:related}
Due to the difficulties of the cell nuclei segmentation task, most algorithms apply a sequence of processing methods. We explain some of these methods, their goals and their relation to each other.

\subsubsection{Graph-based Image Segmentation}
A straightforward approach for graph-based image segmentation is to define edges with weights based on the similarity between neighboring pixels and then minimize the cut (maximize dissimilarity) between two regions. However, this approach is biased towards unnaturally small regions, because these have less outgoing edges and hence often a smaller cut. Therefore, a second important graph-based image segmentation technique is the minimization of the \emph{normalized cut} \cite{shi00normalized}. This criterion maximizes both the similarity inside a region and the dissimilarity between regions. 

Another popular method is the graph cut framework (see Section~\ref{sec:pre:graphcuts}). \citet{alKofahi10improved} apply it both for image binarization and splitting of nuclei clusters. 
In recent experimental evaluations, graph cut-based cell nuclei segmentation methods have ranked best regarding the segmentation quality \cite{stegmaier14fast}.

The graph cut multi-label techniques' running times grow linearly with the number of labels, i.e. the number of image objects. This drawback makes additional work necessary to render them practicable for more than 20 labels \cite{alKofahi10improved}.

\subsubsection{Binarization} Often a first step in image analysis is a binarization of the image, with the aim to distinguish the foreground from the background.

Image binarization algorithms such as Otsu's method \cite{otsu75threshold} or adaptive iterative thresholding \cite{keller08reconstruction} can work well if the cells are isolated and foreground and background intensity ranges are sufficiently far apart. Combined with morphological filtering, they enable a fast binarization \cite{lin03hybrid}.

Graph cut based binarization in the context of cell nuclei has been applied several times.
\citet{danek12thesis} heuristically determines hard constraints for the data term, the boundary term employs a sophisticated computation of the distance between pixels in a gradient induced Riemannian manifold. This technique has high computational demands.
\citet{alKofahi10improved} perform minimum error thresholding to gain probability distributions for the data term, and use a simple gradient-based smoothness term.

If the cells are isolated, a single image binarization can be used to successfully segment the nuclei.
However, for touching or overlapping objects, the task is more complex. The binarization step computes an \emph{under-segmentation} of the image, resulting in clusters of cell nuclei. These have to be split in a cluster separation step.

\subsubsection{Seed Detection}
In terms of image analysis, detection refers to the task of placing a marker, or a seed point, on each desired object in the image.
A basic insight of image analysis is that detection is easier than segmenting, i.e. delineating the border of objects. Therefore, an actual segmentation is often preceded by a seed detection phase. This especially applies to the field of cell nuclei segmentation, where the desired objects are roundish \emph{blobs} of similar sizes.

The popular multiscale Laplacian of Gaussian (LoG) provides an effective blob detector. \citet{stegmaier14fast} find peaks in the maximum intensity projection of a range of predefined scales, while \citet{alKofahi10improved} implement a computationally more expensive automatic scale detection, exploiting cues in the distance transform of the binary foreground mask. \citet{danek10oneuclidean} search for peaks in the distance transform. While this method is fast, it is error prone for tightly clustered nuclei, especially in 3D. \citet{lou12learning} use a blob detector based on eigenvalues of the Hessian matrix.

The success rate of later stages depend on the accuracy of the seeds, as each seed will later form a final object. 

\subsubsection{Cluster Separation}
When faced with an under-segmentation, object clusters have to be split to distinguish each cell nuclei.
The classical approach is the watershed algorithm. However, it often results in over-segmentation, making a post-processing step necessary \cite{lin03hybrid}. 
The graph cut framework with its multi-label variant has been used as well. \citet{danek10oneuclidean} use the above described seed points as initial labels and combine the boundary term from the binarization phase with a distance transform induced shape term.
In \cite{alKofahi10improved}, the data term is modeled with a Gaussian mixture for each cell, and the ad hoc smoothness term is reused. The $\alpha$-expansion technique is sped up by reducing the number of labels. Instead of introducing one label per seed, the labels are found by coloring a graph which is built on the initial labels.
In \cite{lou12learning} the probability distribution is gained with supervised learning, by training a random forest on a set of local features. This approach requires a set of images with ground truth -- a fact which makes this method difficult to employ for huge 3D images with thousands of cell nuclei. The authors also introduce shape priors: they augment the graph cut formulation with a term which encourages cuts to be aligned to a vector field centered at the initial seeds.

\citet{he15icut} employ the normalized cut algorithm \cite{shi00normalized} to segment 2D cell nuclei images. The image is represented with a fully connected graph. The edge weights encode gradient information, and weights of edges which cross binarization contours are set to zero. The optimal normalized cut is computed by solving an eigenvector system. Clusters consisting of more than two nuclei are split recursively, with a heuristically fixed number of recursions. Depending on this number, the image might be over- or under-segmented.
The complexity of solving the eigenvector system is $\bigO(n^3)$, where $n$ is the number of image pixels. It is unclear if this method can be extended to big 3D images.

\subsubsection{Cluster Merging}
Instead of performing a top-down approach, where an under-segmentation is split up in a postprocessing step, it is also possible to segment the image in a bottom-up fashion. The idea is to use an algorithm which results in an over-segmentation and then merge the components to intact nuclei.
\citet{lin03hybrid} apply a 3D watershed algorithm on a gradient-weighted distance transform, which allows them to incorporate both geometric and intensity cues.
In a postprocessing step, they recursively merge touching objects. 
Merging blocks of pixels is conceptually simpler than splitting them, because the cut is already defined. The difficulty is to determine which segments are to be merged. In \cite{lin03hybrid}, a score is computed which signifies the coherence of the potential merged objects to a nuclei model. The highest score determines which objects are merged next. However, future potential merging steps are not considered. An improvement \citet{lin05hierarchical} is to make merge decisions by finding the best solution in merge trees built on the nodes of a region adjacency graph. This graph encodes all possible sequences of merging steps.
\citet{liu01region} combine splitting and merging of objects in one algorithm. The algorithm recursively splits regions by straight lines between candidate endpoints.

\subsubsection{Prior Knowledge and Model-driven Approaches}
Best segmentation results cannot be obtained with an all-purpose algorithm. The assumption that the information required for a perfect segmentation is provided by the image only is wrong. Hence, segmentation algorithms have to incorporate domain knowledge to be successful. This knowledge can be of qualitative or quantitative nature. In the context of nuclei segmentation, for example, the qualitative knowledge that cell nuclei are blob-shaped is exploited by using a LoG seed detector. Quantitative knowledge is introduced by making use of known values for certain object-specific features. In \citet{stegmaier14fast}, for example, the multiscale LoG-filter works on a range of user-defined scales, which reflects the expected diameter range of the nuclei.

The knowledge on a feature or a set of features can be expressed in a probabilistic \emph{object model}. This model can guide a segmentation process by reflecting the probability that an image object is a desired object or not. Instead of predefining the model parameters, it is also possible to learn them from the data. The merge decision process in \cite{lin03hybrid, lin05hierarchical}, for example, makes use of a model built on eight features, including texture, volume and several quantifiable shape features. The parameters of a multi-dimensional Gaussian distribution over these features are estimated from heuristically chosen objects in the image. This distribution enables the computation of a confidence score for any object. 
\citet{liu01region} 

The split and merge algorithm in \cite{liu01region} is model-driven as well, the parameters for the three features color, shape and area overlap are gained by supervised learning.

% --------------- % 
%  Preliminaries  %  
%---------------- %
\subsection{Preliminiaries}
\label{sec:preliminaries}
\subsubsection{Images}
This section introduces a set of terms and their relation to each other. For a deeper explanation of these terms, we refer to \cite{sonka14image, gonzalez07image, szeliski10computer}.
A \emph{scene} is a segment of the real world. It is composed of three-dimensional \emph{objects}. A \emph{real image} is a continuous function 
$f: \mathbb{R}^2 \rightarrow \mathbb{R}.$ It maps points from the two-dimensional \emph{image plane} to the \emph{intensity}. The 2D real image is the result of a projection of the scene onto the plane \cite{sonka14image}.
To be processed by a computer, the real image is \emph{digitized}. 
A \emph{digital image} $\imagesymbol$ is a function
from a bounded discrete rectangle $\pixelspace = \{0, 1, \dotsc, S_x\} \times \{0, 1, \dotsc, S_y\}$ to the discrete brightness levels $V = \{0, 1, \dotsc, K\}$. An element $p = (i, j) \in \pixelspace$ is called a \emph{pixel}. The rectangle $\pixelspace$ is called the \emph{pixel space}. For the value of a pixel $p$, we also write $\image{p}$ instead of $\imagesymbol(p)$.

The digitization is the conversion of the real image to the digital image. The discretization of the plane is done by \emph{sampling} with a discrete set of sampling points in the plane. \emph{Quantization} splits the intensity range into $K$ intervals. Usually, $K$ is of the form $2^b$, where $b$ is the \emph{bit depth}.
The \emph{histogram} $h: V \rightarrow \mathbb{N}$ of a digital image is the absolute frequency of discrete brightness levels in an image.

Digital images formed by the described process are two-dimensional, because the image sensor used in the digitization is two-dimensional. Images of higher dimensions are formed by grouping a set of images to an \emph{image stack}: a three dimensional image is a stack of two-dimensional images. Multiple 3D images which represent different points in time can be stacked to a 4D (or 3D+t) image.
This work, only deals with 3D images. In this case, the pixel space is a discrete cuboid:
$$\pixelspace = \{0, 1, \dotsc, S_x - 1\} \times \{0, 1, \dotsc, S_y - 1\} \times \{0, 1, \dotsc, S_z - 1\}.$$ A 3D pixel might also be called \emph{voxel}. The \emph{size}~$S$ of an three-dimensional image is the vector~$(S_x, S_y, S_z)$. In the following, when we speak about images, we will refer to 3D digital images.

There are two notions of distances in digital images: \emph{physical distance} or \emph{pixel spacing}, and \emph{pixel distance}.
To be able to measure the physical distance between two real image features using the digital image, the pixel spacing has the be known. It is defined as the shortest distance between two real-world points which can be distinguished with the combination of the optical system and the employed camera, along each of the three axes.
Given the image spacing $\delta \in \mathbb{R}^3$, $\delta = (\delta_x, \delta_y, \delta_z)$, the image \emph{resolution} denotes the pixel per unit distance $r=(r_x, r_y, r_z)=(1/\delta_x, 1/\delta_y, 1/\delta_z)$. Then the physical distance $\dist(p, q)$ between two pixels $p, q \in \pixelspace$ is computed by interpreting $\pixelspace$ as the euclidean three-dimensional space and determining the euclidean distance between $p$ and $q$.

The pixel distance is a dimensionless measure for the amount of basic steps required to move between discrete pixels. We define two variants. The distance $D_6$ is the 3D extension of the city-block distance and is defined as 
$$D_6((x_0, y_0, z_0), (x_1, y_1, z_1)) = \abs{x_1 - x_0} + \abs{y_1 - y_0} + \abs{z_1 - z_0}.$$ 
$D_{26}$ is known as the checker-board distance:
$$D_{26}((x_0, y_0, z_0), (x_1, y_1, z_1)) = \max(\abs{x_1 - x_0}, \abs{y_1 - y_0}, \abs{z_1 - z_0}).$$

Two pixels $p$, $p$ are \emph{$k$-adjacent}, $k \in \{6, 26\}$, if $D_k(p, q) = 1$. Then, $p$ is a $k$-neighbor of $q$. The \emph{$k$-neighborhood} of a pixel is the set of pixels which are $k$-adjacent to it. When clear from the context, or not necessary, we omit the specifier in the following. A \emph{neighborhood system} $\neighborhoodsystem$ is the set of adjacent pairs of pixels.

A \emph{path} from a pixel $p$ to a pixel $q$ is a sequence of pixels $A_1, A_2, \dotsc, A_n$ where $p = A_1$, $q = A_n$ and $A_{i+1}$ is a neighbor of $A_i$ for $1 \leq i < n$. Two pixels are \emph{connected} if there exists a path between them. A \emph{connected component} or simply \emph{component} is a set of pixel which are pairwise connected. The \emph{border} of a component $\component{}$ is the set of pixels of the component with one or more neighbors outside of $\component{}$.
%A \emph{partitioning} $\partitioning{}$ of a component $\component{}$ is a family of (nonempty) components over the pixels of $\component{}$ such that each pixel of $\component{}$ is a contained in at most one $\component{i} \in \partitioning{}$.

A \emph{segmentation family} of an image is\footnote{
Most definitions\cite{gonzalez07image} further demand that $Q(R_i) = \mathtt{TRUE}$ and $Q(R_i \cup R_j) = \mathtt{FALSE}$
for a logical, domain-specific predicate $Q$. While this is certainly true for all segmentations, it is rather vague not necessary for a definition. 
} a finite set of subregions $R_1, \dotsc, R_l$, where
\begin{itemize}
\item $R_i$ is a connected component, 
\item $\pixelspace = \bigcup_{i=1}^l R_{i}$, and
\item $R_i \cap R_j = \emptyset$ for $i \neq j$.
\end{itemize}
It is possible to relax the second demand by introducing a special not necessarily connected subregion $R_b$ for the background defined as $R_b := R \setminus \bigcup_{i=1}^l R_{i}$.

For a given segmentation family, a \emph{segmentation} is a function which maps pixels to the subregions they are contained in. The subregions $R_i$ are then also called \emph{labels}.
A special case of a segmentation is a \emph{binarization}, where the number of subregions is $l = 2$, and the regions are foreground and background, denoted as $R_f$ and $R_b$, respectively.

Informally, each component of a segmentation corresponds to a real world object in the scene represented by the digital image. Hence, each pixel is mapped to either a real world object, or background.
Another notion of segmentation is \emph{soft segmentation}, which takes into account that a particular pixel can incorporate signals of multiple objects. This is accomplished by making use of fuzzy set theory \cite{udupa96fuzzy}. Each pixel is mapped to a fuzzy set over $R_1, \dotsc, R_s$. This technique should not be confused with the uncertainty propagation discussed in the next section, where fuzzy sets are used to measure the uncertainty of the entire (hard) segmentation. We only consider hard segmentation in our work.

\subsubsection{Uncertainty Evaluation}
\label{sec:uncertainty}
Most operators in an image processing pipeline are unable to give precise results for all inputs. On the one hand, this is due to the inherent deficiency of the imaging quality. The low signal-to-noise ratio might it make hard to distinguish small nuclei from background, or aberrations
might be wrongly detected as nuclei. On the other hand, many tasks in image processing are ill-posed problems \cite{khairy11reconstructing, danek12thesis}: the solution to such a problem are subject to ambiguities. There might not be exactly one correct answer regarding a certain image feature, e.g. a cell which is in the mitosis during the imaging process can be detected as a single cell or two cells, depending on how far the mitosis is already visible. 
In order to prevent errors made by one operator from affecting downstream tasks, but also to avert the loss of improbable but viable solutions, a stage should be able to inform its successors of the uncertainty involved in its output. This is made possible by introducing a measure for the validity for every extracted piece of information and propagating it together with the results to the next pipeline stage \cite{stegmaier12challenges}. 

We follow \citet{stegmaier12challenges} in making use of fuzzy set theory to measure uncertainty, based on \emph{a priori} knowledge. They introduce a fuzzy set membership function $\mu_{ijm} \colon \mathbb{R} \to [0,1]$ for every operator $i$, feature $m$ and linguistic term $j$. 
One way to define such a function is by using the following set of four terms: ``Feature is\dots $\{$\dots as expected'';\dots too small, but useful'';\dots too large, but useful'';\dots not useful''$\}$.
This set lets us model the uncertainty with a trapezoidal membership function $\mu_{m}(x, \theta_m)$ defined by a quadruple $\theta_m = (a, b, c, d)$ as\\
\begin{equation*}
\label{eq:trapezoidal_membership}
    \mu_{m}(x, \theta_m) = 
    \begin{cases}
		\frac{x - a}{b-a}	& \text{if $a \leq x < b$;}\\
		1      				& \text{if $b \leq x < c$;}\\
		\frac{d - x}{d-c}	& \text{if $c \leq x < d$;}\\
		0               	& \text{otherwise.}\\
    \end{cases}
\end{equation*}
Following the rules of fuzzy set theory, multiple uncorrelated features can be combined by multiplying their respective membership functions.

\subsubsection{Balanced Graph Partitioning}
Given an undirected graph $G = (V,E)$ with non-negative edge weights $c: E \longrightarrow \mathbb{R}^+$, and a number $k \in \mathbb{N}$, $k > 1$, the \emph{graph partitioning problem} is to find $k$ sets of nodes $V_1, \dots, V_k$, with $\bigcup_{i=1}^k V_i = V$ and $V_i \cap V_j = \emptyset\;\forall i \neq j$.
Given an additional imbalance factor $\varepsilon \in \mathbb{R}^{+}$, the \emph{balanced graph partitioning problem} further demands that the sets are balanced, i.e. that $|V_i| \leq (1+\varepsilon) \lceil|V|/k\rceil$. The goal is to minimize an objective function. In this work, we will aim to minimize the sum of the weights of the set of cut edges $\left\{ \left\{ u, v \right\} \in E \mid u \in V_i, v \in V_j, i \neq j \right\}$.
The graph partitioning problem has been shown to be NP-complete \cite{hyafil73graph, garey74some}. Practical graph partitioning tools rely on heuristics to find good partitions. Most tools use the multilevel approach \cite{hendrickson95multilevel}, where the graph is first iteratively coarsened, i.e. a hierarchy of graphs is constructed which keeps the original structure but reduces the input size. A common method for coarsening is contracting edges in a matching in $G$ \cite{hendrickson95multilevel}.
Then a partitioning is computed on the coarsened graph. One way to do this is to search for $k$ seed nodes which are far away from each other and then alternately run $k$ BFS, one from each seed \cite{diekmann00shapeoptimized}.
Finally the graph with its partitioning is unpacked, while employing local search methods, such as max-flow min-cut based search \citep{sanders11engineering}
and FM \cite{fiduccia82lineartime}. As we cannot cover all methods here, we refer the reader to \cite{bichot13graph} for a deeper insight. 
From the many available tools and libraries for graph partitioning, we tested METIS \cite{karypis98fast} and KaHIP \citep{sanders12think}, and achieved best results with a fine-tuned configuration for KaHIP.

compute exact solutions to the two-label segmentation problem, i.e. image binarization. 

\subsubsection{Graph Cut Segmentation}
\label{sec:pre:graphcuts}
The graph cut framework is a powerful graph-based segmentation method.
It has been shown \cite{greig89exact, boykov01fastapproximate, boykov06graphcuts} that it is possible to efficiently minimize an \emph{energy function} of the form $E(\segmentationsymbol)$ for a binarization $\segmentationsymbol$ defined as
$$E(\segmentationsymbol) = \lambda E_{\mathrm{data}}(\segmentationsymbol) + E_{\mathrm{smooth}}(\segmentationsymbol).$$
A minimum of this function corresponds to an exact solution to the two-label segmentation problem, i.e. image binarization.

The data term is typically
$$E_{\mathrm{data}}(\segmentationsymbol) = \sum_{p \in \pixelspace} D_p(\segmentation{p}),$$
where $D_p$ measures the pixel-wise penalty of mapping pixel $p$ to label $\segmentation{p}$, i.e. fore- or background. The smoothness term $E_{\mathrm{smooth}}(\segmentationsymbol)$ penalizes neighboring pixels with different labels. Using a neighborhood system $\neighborhoodsystem$, it is defined as
$$E_{\mathrm{smooth}}(\segmentationsymbol) = \sum_{(p, q) \in \neighborhoodsystem} B_{p, q} \cdot 1_{\segmentation{p} \neq \segmentation{q}}.$$
Here, $1_{\segmentation{p} \neq \segmentation{q}}$ is an indicator function which is $1$ iff $\segmentation{p} \neq \segmentation{q}$. The function $B_{p, q}$ is the boundary energy.
The energy function $E(\segmentationsymbol)$ is minimized by computing a minimum cut on a special graph. 

\citet{boykov01fastapproximate} extended the method to multi-label energy minimization, which is NP-hard in general. They find approximate solutions by introducing two approximation techniques, $\alpha$-expansion, and $\alpha/\beta$-swap. 
The former performs iterative binary segmentations between one label and every other label. The latter computes the minimum cut between all pairs of labels.

The graph cut segmentation method is only a framework. To employ it in practice, the terms $D_p$ and $B_{p, q}$, and the scalar $\lambda$ have to be parameterized. \citet{boykov06graphcuts} suggest to determine models for the pixel intensity distributions of the objects to be segmented. The data term can then be parameterized using the fit of the pixel value to the models. Hard constraints can be incorporated as well.

Due to the extensive use of the graph cut framework for segmentation problems, maximum flow algorithms which exploit the special structure of the graph have been developed \cite{boykov04experimental, goldberg11maximum} and parallelized \cite{delong08scalable, liu10parallel, jamriska12cacheefficient}.

\subsubsection{Cut Metrics}
Due to the discrete nature of digital images, computing the surface area of a real-world object using a digital 3D image\footnote{or the length of a contour in 2D} is a nontrivial problem. Interpreting each voxel as a cuboid and summing up the area of their surfaces is only a bad approximation, since with that method, every object has the same surface area as its bounding box. 
In the context of adopting graph cut to compute geodesics in Riemannian spaces, \citet{boykov03computing} have introduced the notion of \emph{cut metrics}. The authors have shown that it is possible to compute edge weights for a regular grid graph such that the length of a contour is approximated by the sum of the edge weights cut by that contour.
Their method is based on the Cauchy-Crofton formula. This formula relates the Euclidean length $\vert\mathcal{C}\vert$ of a curve $\mathcal{C}$ to the function $n_\mathcal{C}(l)$ which measures the number of intersections between the curve $\mathcal{C}$ and a line $l \in \mathcal{L}$:
\begin{equation}
\vert\mathcal{C}\vert = \frac{1}{2} \int_\mathcal{L} \! n_\mathcal{C}(l) \, \mathrm{d}l = \frac{1}{2} \int_0^\pi \int_{-\infty}^{+\infty} \! n_\mathcal{C}(\phi, \rho) \, \mathrm{d}\phi\,\mathrm{d}\rho, 
\end{equation}
using that the space of lines can be represented as $[0, \pi] \times [-\infty, +\infty]$.
\citet{boykov03computing} approximate the integral with partial sums by partitioning the space of lines among the edges of the grid graph's neighborhood system.
This method yields edge weights 
\begin{equation}
\omega_k = \frac{\delta\phi_k \delta\rho_k}{2},
\end{equation}
where $\delta\phi_k$ is the angular partition of edge $k$ and $\delta\rho_k$ is the distance of the line induced by edge $k$ of the neighborhood system to the nearest parallel edge induced line.

The method can be formulated for 3D grids as well. However, for that case \citet{boykov03computing} do not specify how to partition the space of angular orientations among the edges of the neighborhood system.
\citet{danek10oneuclidean} proposed a solution both in 2D and 3D, as well as for grids with anisotropic node spacing. They compute the Voronoi diagram of the intersections of the neighborhood system with a unit hypersphere. The angular partitioning of an edge is equal to the fraction of its Voronoi cell.

\section{Our Method}
\label{sec:methods}
This section deals with our segmentation method. Our method performs a top-down approach. We first perform a fast and robust foreground detection, detailed in Section~\ref{sec:binarization}. The connected components in the binarized image are then recursively split. We make use of balanced graph partitioning to cope with the bias towards small partitions when minimizing a cut (cf.  \@Section~\ref{sec:partitioning}). Our segmentation framework computes a score for each component which is based on prior knowledge. The nucleus model is presented in Section~\ref{sec:model} and the segmentation framework Section~\ref{sec:framework}. The advantage of our cluster segmentation approach over the approaches described in Section~\ref{sec:related} is that our method does not depend on a heuristic seed detection, which improves the detection and segmentation quality for highly clustered cell nuclei.

% --------------- % 
%  Binarization   %  
%---------------- %
\subsection{Binarization}
\label{sec:binarization}
The first step of our segmentation algorithm is to binarize the image. We aim to classify the image pixels into two groups, the foreground and the background. This step immediately reduces the input, as all background pixels do not have to be further examined. 
For our purpose, it is important that foreground objects not be 
wrongly classified as background. An error of this kind leads to undetected nuclei because our subsequent steps are not capable of detecting these objects. Another binarization mistake are non-foreground areas which are wrongly detected as foreground. This case might arise due to defective camera sensors, e.g. isolated white pixels. To some extend, our splitting framework is able to detect and discard these erroneous regions especially in the aforementioned case.

A third binarization error is an inexact delineation of the foreground object's border. 
The object is correctly detected as foreground, but its shape or size differs. 
This error commonly arises in binarized images of confocal fluorescence microscopy. 
On the one hand, unstained parts of a nucleus may appear as dark as the background \cite{svoboda09generation}. 
On the other hand, an enlargement of the nuclei can be observed for binarized cell nuclei images. 
The diameter of binarized nuclei is several pixels larger than the real extent of the nuclei. 
The reason for the inflated objects is the optical diffraction, which can be expressed by the \emph{point spread function} (PSF) of the imaging system. The PSF represents the response of a pixel sensor to an ideal point light source \cite{szeliski10computer}.
A point source appears blurred in the image and might be surrounded by concentric rings. 
The area directly adjacent to the foreground objects is then still brighter than the dark background. 
This effect can add up if several objects are close to each other.
In these cases, clustering-based thresholding approaches which do not use domain knowledge result in inflated objects. 

This phenomenon effects our nuclei segmentation method both in quality and in running time. Firstly, the quality suffers. Our method relies on splitting the foreground regions of the image. It only adds cuts inside the foreground, but does not change the delineation between fore- and background. Furthermore, the inflation of the nuclei can also cause well-separated objects to be connected in the binarization. These connecting bridges will then be part of one the segmented objects. These circumstances would have a negative impact on the quality of the segmentation results. 
Secondly, the segmentation speed would be slowed down. Our method uses graph partitioning to split the foreground objects. The larger number of merged objects requires more partitioning steps, and their inflated size results in overall bigger graphs to be partitioned.

The PSF depends on the imaging system. It is a combination of the blur induced by the optical system and the finite
integration area of a chip sensor \cite{szeliski10computer}. Hence, the extend of the foreground blur can differ across distinct datasets. In one dataset, the blur might be strong, and measures have to be taken to detect and cope with it. In another set, simple thresholding already leads to good results. In this work, we experimented with three different binarization methods. The first is Otsu's method \cite{otsu75threshold}, a simple thresholding algorithm. The other two are based on work by \citet{restif06segmentation}, who models the gray level histogram of a blurred image. The model is described in the next section. We either use this model to compute a threshold, or parameterize the graph cut framework with its probability distributions.

For all algorithms, binarization quality could be improved by convolving the image with a low-pass Gaussian filter with small standard deviation $\sigma_s$. This method reduces background noise and eliminates one-pixel errors.
Another image degradation specific to confocal microscopy is uneven illumination, a brightness gradient in the image, often along the axial direction \cite{muller06introduction}.  We overcome this deficiency by dividing the image into sets of consecutive slices of size $m$ and binarizing each set separately.

% --------------- %
%  Image model    %  
%---------------- %
\subsubsection{Image Histogram Model}
\label{sec:histogram_model}
We compute the normalized gray level histogram $h(i)$ with the goal to infer two distributions, one for the foreground and one for the background. \citet{alKofahi10improved} assumed both to be Poisson distributed, citing the image formation process and their findings of the histograms to be bimodal.
However, for some of our data we observed unimodal histograms, due to the small fraction of foreground compared to the entire image, the high intensity variance in the fluorescing nuclei, and the system-specific point spread function, which blurs the image and results in actual background regions to be brighter around the foreground objects.
We therefore adopt a model inspired by \cite{restif06towards}. The background is subdivided into non-illuminated (NB) and illuminated background (IB). The former is a normal distribution with parameters $\mu_b$ and $\sigma_b$, and a priori probability $p_b$:
\begin{equation}
\label{eq:nonilluminated_background}
NB(i) = \frac{p_b}{\sqrt{2\pi}\sigma_b} \exp{\left(-\frac{(i-\mu_b)^2}{2\sigma_b^2}\right)}.
\end{equation}
The IB accounts for the fluorescence microscopy-inherent blurring effect: an increased gray level intensity of the background around the fluorescing nuclei. This blur is modeled with an exponential decline, from which the model for the IB is derived (see \cite{restif06towards}) as
\begin{equation}
\label{eq:illuminated_background}
IB(i) = \frac{2\alpha A}{i - \mu_b}\log\frac{(I_f - \mu_b)}{(i - \mu_b)},
\end{equation}
where $I_f$ is the intensity at the border of the object and $A$ is the object's area. The factor $\alpha$ defines the blur's rate of decline. \citet{restif06towards} introduces a foreground normal distribution and an IB term for each nuclei, which is practical as the images used in his work on average only contain about seven foreground objects. For our purpose, we model the nuclei with one normal distribution $F(i)$ with parameters $\mu_f$, $\sigma_f$, and $p_f$, and the entire illuminated background with one term $IB(i)$. We set $I_f$ to $\mu_f$, and $A$ to $p_f$, the foreground's a priori probability. Because the IB encroaches into the NIB, we only compute the $IB(i)$ in the range $[\mu_b + 2\sigma_b, \mu_f)$.
Then, the histogram model is 
\begin{align}
\label{eq:restif_model}
h_{model}(i) =& NB(i) + IB(i) + F(i)\\
		=& \frac{p_b}{\sqrt{2\pi}\sigma_b} \exp{-\frac{(i-\mu_b)^2}{2\sigma_b^2}} + \frac{2\alpha p_f}{i - \mu_b}\log\frac{(\mu_f - \mu_b)}{(i - \mu_b)} + \frac{p_f}{\sqrt{2\pi}\sigma_f} \exp{-\frac{(i-\mu_f)^2}{2\sigma_f^2}}
\end{align}
with the set of missing parameters $\{p_b, \mu_b, \sigma_b, p_f, \mu_f, \sigma_f, \alpha\}$. 

Like \citet{restif06towards}, we determine the parameters with the expectation--maximization (EM) algorithm \cite{dempster77maximum}. 
The EM algorithm iterates between an expectation step (E-step) and a maximization step (M-step). The former computes weights which define the proportion of each summand in Equation~\ref{eq:restif_model}, given an estimate of the parameters. The latter estimates the parameters, given a set of weights.

We introduce weights $\omega_i^{nb}, \omega_i^{ib}, \omega_i^f$ for the non-illuminated background, illuminated background, and foreground, respectively.
In the E-step, we use Equation~\ref{eq:restif_model} to compute the weights for each histogram bin $i$, e.g.
\begin{equation}
\omega_i^b = NB(i) / h_{model}(i)
\end{equation}

In the M-step, the parameters for the normal distributions are estimated based on the weighted histogram:
\begin{align}
\label{eq:m_step_normals}
p_b =& \sum_{i} \omega_i^b h(i),\\
\mu_b =& \sum_{i} i \omega_i^b h(i),\quad\text{and}\\
\sigma_b =& \sqrt{\sum_{i} (i-\mu_b)^2 \omega_i^b h(i)},
\end{align}
for the background distribution and equivalently for the foreground distribution.
The parameter $\alpha$ is subsequently determined, analogous to \cite{restif06towards}, as 
\begin{equation}
\alpha = \frac{1}{\log \varepsilon} \sqrt{\frac{1}{p_f} \sum_{i = \mu_b + 2\sigma_b}^{\mu_f - 1} \omega_i^{ib} h(i)},
\end{equation}
where $\varepsilon = 2 \sigma_b / (\mu_f - \mu_b)$ is a correction factor for the truncation of the illuminated background model.

Alternating E-step and M-steps, the algorithm converges after a small number of iterations (about 5 in our settings).
We found the outcome of the EM algorithm to be robust and largely independent on the initialization. Therefore, we initialize the parameters with a simple iterative thresholding algorithm \cite{ridler78picture}, where a threshold is defined as the histogram mean, and a new threshold is iteratively computed as the average between the means of the histogram below and above the old threshold. We use the final threshold to compute the means and a priori probabilities of the foreground and background distribution, and set the variances to 1 and $\alpha$ to 0.01.

For subsequent computations which require probability distributions for foreground and background pixels, we will combine the illuminated and non-illuminated background into one term $B(i)$:
\begin{equation}
\label{eq:background}
B(i) = NB(i) + B(i).
\end{equation}
Hence, $B(i)$ ($F(i)$) can be interpreted as the probability that a pixel in a background (foreground) area has gray level $i$.

We implemented two binarization methods using the above distributions. The first is to compute an intensity threshold $t^\star$ as 
\begin{equation}
t^\star = \min\limits_{i \in V} F(i) < B(i).
\end{equation}
All pixels with values above this threshold are determined as foreground, all others are background.

The second method uses these distributions to parameterize the graph cut framework. With the notations in Section~\ref{sec:pre:graphcuts}, we set the pixel-wise data term to
\begin{equation}
D_p(\segmentation{p}) = 
    \begin{cases}
		-\log F(\image{p})	& \text{if $\segmentation{p} = R_f$;}\\
		-\log B(\image{p})  & \text{if $\segmentation{p} = R_b$.}\\
    \end{cases}
\end{equation}
For the smoothness term, we use
\begin{equation}
\label{eq:smoothness_gauss}
B_{p,q} := \exp\left(-\frac{(\image{p} - \image{q})^2}{2\sigma_{\mathrm{bin}}^2}\right) \frac{1}{\dist(p, q)}.
\end{equation} 
This function was suggested in \cite{boykov06graphcuts} and also adopted in \cite{alKofahi10improved}. However, it introduces a parameter, $\sigma_{\mathrm{bin}}$. It results in a high penalty for discontinuities when the intensity difference is below $\sigma_{\mathrm{bin}}$. A the function resembles a Gaussian probability distribution, the parameter can be interpreted as the standard deviation of the image noise. This parameter, and the scalar parameter $\lambda$, have to be set by hand.

% -------------------- % 
%  Graph partitioning  %  
%--------------------- %
\subsection{Cluster Separation}
\label{sec:partitioning}
The binarized foreground in cell nuclei images decomposes into a large number of small connected components. These can either form single nuclei, clusters of nuclei or falsely detected regions due to imaging defects. We perform graph-based segmentation successively on each component.

We use balanced graph partitioning to split the binarized foreground. The graph partitioning aims to minimize the cut between partitions, i.e. the sum of weights of the cut edges. This allows the method to take advantage of intensity or gradient information of the original image, but also of shape cues of the binary foreground mask.

In this perspective, the approach is related to the multi-label cut methods from the graph cut framework, namely $\alpha$-expansion, and $\alpha/\beta$-swap. However, these methods rely on a proper parameterization to show good performance. The data term of the labeling energy graph cut framework reflects the attractions of a voxel to all labels. These have to be determined in advance. With graph partitioning, the number and, more importantly, the positions of the labels do not have to be known. Another difference is that the graph cut methods perform multiple iterations of optimization steps. The $\alpha$-expansion, for example, iterates over all partitions and computes a minimum cut to the rest of the graph. It repeats this process multiple times until convergence. In our method, each graph partitioner call is only done once, and its outcome is either kept or discarded.

\begin{figure}[htbp]
  \centering
  \subfloat[Minimum cut bipartition of the component.]{\includegraphics[width=0.48\textwidth]{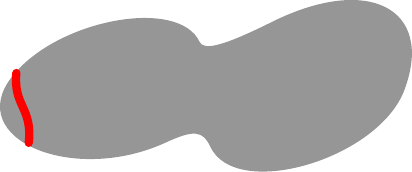}\label{fig:cluster2-1}}
  \hfill
  \subfloat[Minimum cut balanced bipartition of the component. To obey the maximum imbalance factor, cuts have to lie in the green area.]{\includegraphics[width=0.48\textwidth]{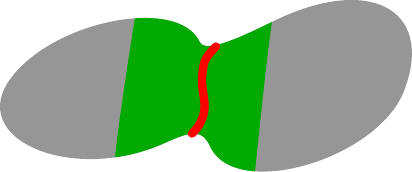}\label{fig:cluster2-2}}
  \caption{Purpose of the balance factor for graph partitioning.}
\end{figure}
The balance constraint serves multiple purposes. Most importantly, it counteracts the bias towards unnaturally small components. Without this measure, the process would almost always cut only a small set of pixels from the main component, as these are connected by a small number of edges. Figure~\ref{fig:cluster2-1} gives an example for this problem. 

The balance constraint forces the graph partitioner to select edges in the green area of Figure~\ref{fig:cluster2-2} and thus finds the cut which correctly segments the image.
This image also illustrates another reason to employ balanced graph partitioning for cell nuclei segmentation. The objects are often roughly equally sized and separated only by tight bottlenecks. Therefore, the combination of minimizing a cut and balancing the two resulting partitions makes adequate use of the shape cues of the binary foreground mask: cuts are smaller when they contain less edges, which favors bottleneck regions to be cut.

The balancing also improves the speed of our algorithm for big components which have to be split multiple times. While the number of required partitioner calls remains asymptotically unchanged, the subcomponents sizes decrease faster, and the recursion depth is decreased.

For each connected component in the binarized image we construct a graph with nodes representing voxels and edges connecting neighboring voxels. A 6-neighborhood was sufficient in our experiments.
As the number $k$ the foreground components have to be split into is not known a priori, we set $k = 2$ and compute recursive bipartitions. The model-based stopping-criterion will be explained in the next section. 

\subsubsection{Edge Weights}
We describe the edge weights for the isotropic case. We implemented two methods of setting the edge weights.
One (\emph{grad}) sets weights based on gradient, the other (\emph{prob}) on the probability of a pixel intensity belonging to background. 

For the gradient-based edge weights, we reuse the method for the graph cut binarization (see Section~\ref{sec:binarization}). The intuition behind this choice is that neighboring pixels with high intensity difference are less likely to belong to same label. 
We compute edge weights $c_{\mathrm{grad}}(\{p, q\})$ between two pixels $p, q$ with intensities $\image{p}$ and $\image{q}$ as
$$c_{\mathtt{grad}}(\{p, q\}) = \exp\left(-\frac{(I_u - I_v)^2}{2\sigma_{\mathrm{grad}}^2}\right).$$
The parameter $\sigma_{\mathrm{grad}}$ is chosen smaller than the $\sigma_{\mathrm{bin}}$ in the binarization.

However, in our case all pixels are already detected as foreground. We experienced problems with this method for images with high variations of the fluorescent marker. The resulting irregular inner-cell texture of the nuclei can lead to wrong cuts if the inner-nuclei gradient is too high.

In contrast to 2D images, tightly clustered cell nuclei in 3D images do not overlap. In our instances, we observed that even between very close or touching nuclei there was still a tight band of darker pixels. Therefore, a second method is to induce the edge weights directly from the gray levels. The cut should tend to be near the darker area between the bright nuclei. In order to be able to quantify the notion of ``dark areas'', we reuse the histogram image model described in Section~\ref{sec:model}.

We maximize the probability that the nodes a found cut $S$ is adjacent to represent background pixels.
To this end, we disregard the non-independence of neighboring pixel values, and aim to maximize the product of the background probabilities of the adjacent pixels. 
Formally, we maximize
$$\prod_{\{u,v\} \in S} \min\{\prob{B|I(u)}, \prob{B|I(v)}\},$$
where we use the conditional probability $\prob{B|I(u)}$ to denote the probability that pixel $u$ with intensity $I(u)$ is in the background.
Maximizing the above expression is equivalent to minimizing
$$\sum_{\{u,v\} \in S} -\log \min\{\prob{B|I(u)}, \prob{B|I(v)}\}.$$
From the last formula, we derive the edge weights $c_{\mathrm{prob}}(\{u, v\})$ as 
\begin{equation}
c_{\mathrm{prob}}(\{u, v\}) = -\log \min\{\prob{B|I(u)}, \prob{B|I(v)}\}.
\end{equation}
The required probabilities that a given intensity represents background can be computed with
\begin{equation}
\prob{B|I(u)} = \frac{p_b \prob{I(u)|B)}}{p_b \prob{I(u)|B} + p_f \prob{I(u)|F}}.
\end{equation}
for our model described in the previous section, this is equal to
\begin{equation}
\prob{B|I(u)} = \frac{B(i)}{h_{model}(i)}.
\end{equation}

For experimental purposes, we also tried setting all edge weights to the same constant. Then the cut is independent of the intensity levels and only depends on the shape of the foreground objects.

\begin{figure}[htbp]
  \centering
  \subfloat[Neither the red nor the green cut obey the imbalance factor for a bipartitioning.]{\includegraphics[width=0.48\textwidth]{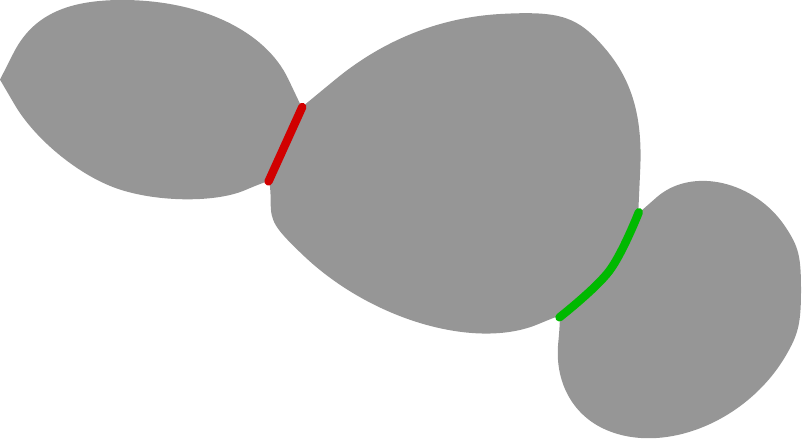}\label{fig:cluster3-1}}
  \hfill
  \subfloat[The minimum cut balanced bipartitioning. One of the two blocks (lilac) is not connected. A BFS results in three subcomponents.]{\includegraphics[width=0.48\textwidth]{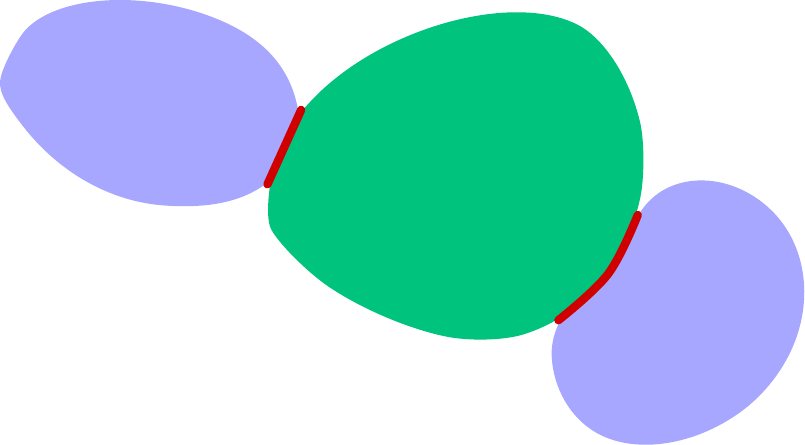}\label{fig:cluster3-2}}
  \caption{Unconnected blocks of the graph partitioning.}
\end{figure}

Once the graph is constructed, we partition it into two blocks with a balanced graph partitioning solver. In general, these blocks do not have to be connected. This circumstance occurs in particular if a component consists of more than two nuclei of differing sizes, as depicted in Figure~\ref{fig:cluster3-1}. Neither the red nor the green cut are eligible. One of the smaller cell nuclei would form one block, and the other block would consist of the big nuclei paired with the other small nuclei, which is not feasible if the imbalance factor is too tight. 
Instead, the balanced graph partitioning results in the green cut shown in Figure~\ref{fig:cluster3-2}. It partitions the graph into two blocks, one (lilac) of which is not connected. 
To cope with this phenomenon, we perform a breadth-first search after the partitioner call to identify the connected components and their sizes.
In the case shown in the example, the component is partitioned into three subcomponents.

In order for the cut in Figure~\ref{fig:cluster3-2} still being valid, even for equally sized nuclei, the balance constraint has to be relaxed. Hence, we set $\varepsilon=0.5$ to cope with this case.

\subsubsection{Anisotropic Resolution} In most confocal fluorescence microscopy images, the resolution along the lateral axes ($r_x$ and $r_y$) is smaller that the axial resolution ($r_z$). When setting the edge weights, we take the anisotropic resolution into account by dividing the weight of an edge between two pixels $u$ and $v$ by their distance $\dist(u, v)$. 

This method is based on the observation that the cut of an object should be invariant of its orientation in the image. The number of edges cut by a 2D plane in the 3D image depends on its orientation. For an example, a cut of the form of an $xy$-plane of area $1$ will cut (i.e. is perpendicular to) $r_xr_y$ edges, while a $xz$-plane of the same size will cut $r_xr_z$ edges. By dividing the edges by their length the sum of these cut edges will be the equal.

% ------------------- %  
%  Scoring function   %  
%-------------------- %
\subsection{Nucleus Model}
\label{sec:model}
We construct a mathematical model of a cell nucleus. The goal is twofold: on the one hand, the model should decide whether an image component represents a single object such that the recursion can be stopped, a cluster of multiple objects to be split, or background noise to be discarded. On the other hand, we want to compute a score $\score{i}$ for each component $\component{i}$ which signifies the probability that it represents a single nucleus.

A splitting step results in a number of subcomponents, each of which is smaller in size than the original component. If a connected component is too big to be a nuclei, we can safely perform a splitting operation. If it is too small, we can be certain that it is not a nuclei. Therefore, in a first approach, we base the scoring function only on the volume $V$ of the components. The volume $V_i$ of an object $\component{i}$ is approximated by multiplying its voxel number $n$ with the spacing:
\begin{equation}
\label{eq:volume}
V_i = n \delta_x \delta_y \delta_z.
\end{equation}
The voxel number $n$ is counted during the breadth-first search. 

From user-defined minimal and maximal cell volumes ($V_{\min}$ and $V_{\max}$, respectively) a trapezoidal fuzzy set membership function $\mu_V$ indicates whether a component has a plausible volume. We define $\mu_V$ with the parameter vector $\theta_V=(V_{\min}, (1+\lambda)V_{\min}, (1-\lambda)V_{\max}, V_{\max})$. %
The parameter $\lambda$ gives us a fine control over the shape of the lateral sides of the trapezoid. In our experiments, it is set to $0.2$.

A component with volume below $V_{min}$ is discarded. A component is repartitioned if
it is big enough to contain a minimum sized nucleus. 
Due to the imbalance factor $\varepsilon$ of balanced graph partitioning, the reduction in size of a subcomponent compared to its original component is at least $(1+\varepsilon)/2$. Therefore, if $V_i$ is greater than $\frac{1+\varepsilon}{2}V_{\min}$, the component $\component{i}$ could be constituted of one minimum sized nucleus $\component{c}$. The remaining part, $\component{i} \setminus \component{c}$, does not have to represent a nucleus.

The volume can be used as a single criterion if the objects' sizes have a small variance. However, we observed that the volume range of cell nuclei allowed both a parent component and its subcomponents to be inside the user-given size limits. 

To improve the decision making process and give better estimates for the uncertainty values, we use the knowledge that our desired objects are cell nuclei, which can be seen as roundish, blob-like objects. 
To distinguish between an ellipsoid-like nucleus and one cut into two nearly equally sized partitions (i.e. similar to an ellipsoidal dome), we compute the sphericity $\Psi_c$ defined \cite{wadell35volume} as the ratio between the surface of a sphere with volume $V_i$ and the surface $A_i$ of the object $\component{i}$.
\begin{equation}
\Psi_i = \frac{\pi^{\frac{1}{3}} (6V_i)^{\frac{2}{3}}}{A_i}.
\end{equation}
where $A_i$ is the surface area of a $\component{i}$. 

In theory, values for the sphericity $\Psi$ are in a range between $0$ and $1$, where $1$ is the sphericity of a perfect sphere. Depending on organism and cell type, different values of sphericity can be expected for cell nuclei in practice. 

We assume an ideal nucleus to be an ellipsoid with near-equal semi-axes and deduced a sphericity of $\Psi_{ideal} = 0.96$ and $\Psi_{min} = 0.81$.
The sphericity membership function is then defined as
\begin{equation}
\label{eq:sphericity_membership}
    \mu_{\Psi}(x) = 
    \begin{cases}
    	0														&	\text{if $x \leq \Psi_{min}$;}\\
		\left(\frac{x - \Psi_{\min}}{\Psi_{\mathrm{ideal}} - \Psi_{min}}\right)^2	& \text{if $\Psi_{min} < x < \Psi_{\mathrm{ideal}}$;}\\
		1      													& \text{if $\Psi_{\mathrm{ideal}} \leq x$;}\\
    \end{cases}
\end{equation}

We compute the surface area $A$ with the cut metric results described in Section~\ref{sec:pre}. The set of cut metric edge weights in a regular grid are the same for each node. 
We use a 26-neighborhood for the cut metric evaluation because the approximation gained with a 6-neighborhood is to inaccurate.

We multiplicatively combine the two membership function values to gain the score $\score{i}$ of a component: $$\score{i} := \mu_V(V_i)\mu_{\Psi}(\Psi_i).$$

% ------------ % korrekturlesen
%  Framework   %  
%------------- %
\subsection{Splitting Framework}
\label{sec:framework}
We recursively split the connected components with balanced graph bipartitioning, computing the above score at each step.
A way to see this splitting process is a depth-first search through a tree, a set of potential desired objects. Each inner node in the tree represents a component. The initial component forms the root of the tree, the children of a node are the outcomes of a partitioning step. The leaves are the desired objects. Listing~\ref{lst:framework} gives a top-level overview of this depth-first search framework.
\begin{algorithm}[htbp]
\caption{The cell partitioning framework.}
\label{lst:framework}
\begin{algorithmic}[1]
\Function{RecursiveSplit}{$\component{}$}{($k$, $\partitioning{}$)} \Comment{$k$: number of components}
\State $\partitioning{} \gets$ \Call{PartitionerCall}{$\component{}$, $k$} \Comment{$\partitioning{}$: partitioning of $\component{}$}
\ForAll{$\component{i} \in \partitioning{}$}
	\State $(\mathrm{decision}, \score{i}) \gets $ \Call{ScoreFunction}{$\component{i}$}
	\If{$\mathrm{decision} = \mathtt{repartition}$}
		\State $(k_i, \partitioning{i}) \gets $ \Call{RecursiveSplit}{$\component{i}$} \Comment{recursive call}
		\If{$k_i = 0$} \Comment{backtrack}
			\If{$\score{i} > 0$}{\ $\mathrm{decision} \gets \mathtt{keep}$}
			\Else{\ $\mathrm{decision} \gets \mathtt{discard}$}
			\EndIf
		\Else
			\State merge $\partitioning{i}$ into $\component{i}$ of $\partitioning{}$
		\EndIf
	\EndIf
	\If{$\mathrm{decision} = \mathtt{keep}$}{\ $k_i \gets 1$}
	\ElsIf{$\mathrm{decision} = \mathtt{discard}$}
		\State remove $\component{i}$ from $\partitioning{}$
		\State $k_i \gets 0$
	\EndIf
\EndFor
\State $k \gets \sum\nolimits_i k_i$
\EndFunction
\end{algorithmic}
\end{algorithm}

The framework's core is a scoring function (\textsc{ScoreFunction}), which determinates the next processing step for a given component. A component can either be kept, discarded, or repartitioned. Keeping a component means that the search is stopped at that point and the component is not partitioned. In the DFS analogy the component is marked as a leaf. A component which is repartitioned might, but does not have to, consist of multiple desired objects. For a discarded component the search is stopped as well, but the component is deleted from the search tree. The scoring function also computes a score for each component.

The score $\score{i}$ measured for a component $\component{i}$ is interpreted as the probability for an event $\event{i}$, that the component $\component{i}$ is a desired object, under the condition that its parent component $\component{p}$ is not a desired object:
\begin{equation}
\score{i} = \prob{\event{i}\:\vert\: \bar{\event{p}}}.
\end{equation}
A component which is kept must have a score value greater than $0$. If both the score $\score{p}$ of a component $\component{p}$ and the score $\score{c}$ of one of its children is strictly positive, we keep the component which has a higher probability to be a nuclei. Using the fact that for the event $\event{c}$ to occur, the event $\event{p}$ cannot occur, we deduce:
\begin{alignat}{4}
&&\prob{\event{c}} &> \prob{\event{p}}\nonumber\\
\Leftrightarrow\mkern40mu &&\prob{\event{c}\:\cap\: \bar{\event{p}}} &> \prob{\event{p}}\nonumber\\
\Leftrightarrow\mkern40mu &&\prob{\event{c}\:\vert\: \bar{\event{p}}} \cdot \prob{\bar{\event{p}}} &> \prob{\event{p}}\nonumber\\
\Leftrightarrow\mkern40mu && \score{c} \cdot (1-\score{p}) &> \score{p}\label{eqn:score_decision}
\end{alignat}
Thus, for each child component of a component $\component{p}$, we check whether it is discarded or not with Equation~\ref{eqn:score_decision}. As an implication, components with a score greater than $0.5$ are not split.

\begin{figure*}[htb]
\centering
\includegraphics[width=\textwidth]{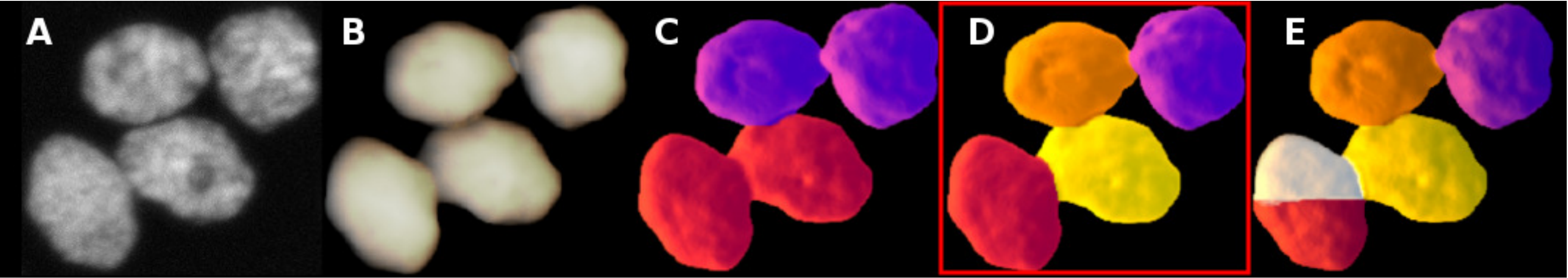}
\caption{An exemplary execution of our pipeline. The input image (A) is first binarized (B), then recursively split (C-E). The split nuclei in the last image (E) is detected by our cell nucleus model. Image D is the final segmentation.}
\label{fig:exemplary_algorithm}
\end{figure*}

We examine each child separately such that discarded components have no influence on their siblings. That way, our algorithm is allowed to discard areas which do not contain any nuclei, reflecting the possible existence of wrongly binarized areas.
In the backtracking step of our depth-first search, we check if any of the child components were kept. Otherwise, the current (parent) component is discarded or kept if its score is greater or equal to $0$, respectively. 
Figure~\ref{fig:exemplary_algorithm} gives an example. The purple nuclei in image D has a low score of $0.46$ due to its upper-ranged size. It is split into two components (E), with scores $0.12$ and $0.05$. After comparing both to the original score, they are discarded during backtracking. Image D is the final segmentation.

\section{Experiments}
\label{sec:experiments}

\subsection{Setup}
We implemented our algorithm in \Cpp using the Insight Toolkit \cite{ibanez05itk} for image operations. The test machine is an Intel Core i7-920, equipped with four cores clocked at 2.67 GHz. The machine has 12 GiB of RAM. All experiments were performed on a single core.
The graphs are partitioned with the open source graph partitioner KaHIP \cite{sanders12think} with a customized configuration.
We compare our algorithm to TWANG \cite{stegmaier14fast} and the graph cut (GC) segmentation described in \cite{alKofahi10improved}. We also include our binarization method (Otsu) in the evaluation.

We measured the quality of a segmentation by counting the number of added, split, merged and missing objects. Added and split are false positives, areas detected as nuclei where there is none. Merged and missing are false negatives. We divide by the number of labeled nuclei in the respective image. The percentages are then averaged over all images in the set. The times measured are averaged as well and do not include image I/O. 
None of the implementations were explicitly optimized for memory consumption. Nevertheless, we display values for memory consumption by extracting the peak resident set size of the respective process.

\subsection{Instances}
Ground truth of real world 3D microscopy images which contain thousands of cell nuclei does not exist. Manual segmentation of these images is not an option due to the large amount of objects on the one hand and the researcher's bias on the other hand. We therefore benchmark the quality of our algorithm on two sets of synthetic cell nuclei images. We downloaded a set of 30 HL60 cell line images generated with the CytoPacq toolbox \cite{svoboda09generation}, with high noise and clustering probability of 75\% (High75). They contain 20 nuclei each and have an isotropic resolution. We scaled the images down by a factor of two in each dimension to let nuclei size resemble those of real world images. Each downscaled image consists of 64 slices with a resolution of $403\times282$.
Based on this toolbox, \cite{stegmaier16generating} have conceived a more realistic set of 3D+t benchmark nuclei images. We use a set of 10 images of different time points (SBDE) which contain between 316 and 1016 nuclei. They have a size of $128\times640\times640$ and the physical distance between two slices is five times greater than the resolution of each slice. 

\subsection{Variants and Parameters}
On the synthetic datasets, the best binarization results were achieved with applying Otsu's thresholding method after convolving the input image with a low-pass Gaussian filter with parameter $\sigma_s$.
This procedure showed robust results on our image sets, even on those without strictly bimodal intensity histograms. Computing a sophisticated image histogram model (Section~\ref{sec:histogram_model}) and using the parameters to threshold the image had worse results. Parameterizing the graph cut framework exhibited a bad trade-off between binarization quality and speed.

For High75, the partitioner edge weights using on the intensity probability ($c_{\mathrm{prob}}$) had slightly better segmentation result than the gradient-based edge weights ($c_{\mathrm{grad}}$). For SBDE however, the situation was inversed, with the edge weights computed by $c_{\mathrm{prob}}$ failing to capture the true object boundaries. The reasons are that the inner-nuclei variance is very low for SBDE, while the nuclei are very tightly clustered, such that there is no ``background area'' between the objects. We therefore use $c_{\mathrm{grad}}$ in our comparison. 
\begin{table}[htbp]
\caption{Parameters for High75 and SBDE, respectively}
\label{tab:parameters}
\centering
\begin{tabular}{l|r|r}
\toprule
Parameter & High75 & SBDE \\ \midrule
$\sigma_s$ 				& 1.9 & 0.7 \\ 
$m$						& 1 & 16 \\		
$\sigma_{\mathrm{grad}}$ 				& 15 & 100 \\
$(V_{min}, V_{max})$ 	& $(20000, 39000)$ & $(2900, 8550)$ \\
\bottomrule
\end{tabular}
\end{table}

Table~\ref{tab:parameters} contains values for the parameters we used in our comparisons. The standard deviation $\sigma_s$ is used for the Gaussian filter preceding the binarization and depends on the size of the objects to detect. The parameter $m$ is the number of regions the image is split into for the binarization and reflects the change of the intensity distribution along the visual axis. The edge weights are parameterized with the parameter $\sigma_{\mathrm{grad}}$, which separates the inner-nucleus intensity variance from the inter-nucleus intensity variance. The minimum and maximum volumes $V_{min}$ and $V_{max}$ are the limits used by our model. The difference in volume between the two image sets stems from the dimensionless nature of the synthetic images.

\subsection{Comparison with other Methods}
% CONNECT sqlite:/home/arz/Documents/Aufschriebe/Ergebnisse/CytoPacq/HL60_3D/HL60_3D.db
\begin{table}[htbp]
\caption{Segmentation results for High75.}
\label{tab:experimentsCytoPacq}
\centering
\begin{tabularx}{\linewidth}{l|rrYYYY}
\toprule
Algo & [s] & [\si{\mebi\byte}] & Added & Missed & Merged & Split \\ \midrule
Otsu & 0.4 &  & 0 & 0 & 71.3 & 0 \\[3pt]
%% TABULAR REFORMAT(col 1-2=(precision=0) col 3-6=(precision=1))
%% SELECT replace(replace(replace(algo, 'alkofahi_nofinalize', 'GC'), 'twangHigh75downscaled', 'TWANG'), 'mine', 'Ours') as _algo, round(avg(ifnull(factor, 1)*(ifnull(bintime, 0) + segmtime))/1000.0, 0) as _time, round(max(ifnull(VmHWM, 0))*1000.0/1024/1024, 0) as _MiB, round(avg(100.0*added/images.numCells), 1) as _added, round(avg(100.0*missing/images.numCells), 1) as _missing, round(avg(100.0*merged/images.numCells), 1) as _merged, round(avg(100.0*split/images.numCells), 1) as _split
%% FROM segmentation
%% JOIN images ON segmentation.dataset = images.dataset and segmentation.num = images.num
%% WHERE machine=112 and segmentation.dataset like "High75downscaled" and (algo like "alkofahi_nofinalize" or algo like "twangHigh75downscaled" or (algo like "mine%" and k=4 and binarization like "RegionalGaussianOtsu"))
%% GROUP BY algo, k, binarization;
   GC & 244 & 174 & 0 & 0 & 0.5 & 1.2 \\
 Ours &   8 & 432 & 0 & 0 & 0 & 0 \\
TWANG &  14 & 251 & 0 & 0 & 0 & 0 \\
% END TABULAR SELECT replace(replace(replace(algo, 'alkofahi_nofinalize', 'GC...
\bottomrule
\end{tabularx}
\end{table}

Table~\ref{tab:experimentsCytoPacq} shows results for the High75 dataset. Both TWANG and our algorithm have no segmentation error on this dataset. GC has a low ratio of merged nuclei and split nuclei. Our algorithm is the fastest. %, only beaten by a four-threaded parallel version of TWANG. 
With the unscaled images, quality is unchanged. GC takes three and a half hours per image, TWANG about 4 minutes and our algorithm less than \SI{90}{\second}, which shows that our algorithm scales best with increasing resolution. This result is due to the fast binarization, which can quickly (\SI{415}{\milli\second}) reduce the amount of voxels to look at by about \SI{91}{\percent}.

% CONNECT sqlite:/home/arz/Documents/Aufschriebe/Ergebnisse/CytoPacq/SBDE/SBDE.db
\begin{table}[htbp]
\caption{Segmentation results for SBDE}
\label{tab:experimentsSBDE}
\centering
\begin{tabularx}{\linewidth}{l|rrYYYY}
\toprule
Algo & [s] & [\si{\gibi\byte}] & Added & Missed & Merged & Split \\ \midrule
Otsu & 4 & & 0 & 0 & 56.2 & 0 \\[3pt]
%% TABULAR REFORMAT(col 1=(precision=0) col 2=(precision=2) col 3-6=(precision=1))
%% SELECT replace(replace(replace(algo, 'alkofahi_nofinalize', 'GC'), 'twangSBDE', 'TWANG'), 'mine', 'Ours') as _algo, round(avg(ifnull(factor, 1)*(ifnull(bintime, 0) + segmtime))/1000.0, 0) as _time, round(max(ifnull(VmHWM, 0))*1000.0/1024/1024/1024, 2) as _GiB, round(avg(100.0*added/images.numCells), 1) as _added, round(avg(100.0*missing/images.numCells), 1) as _missing, round(avg(100.0*merged/images.numCells), 1) as _merged, round(avg(100.0*split/images.numCells), 1) as _split
%% FROM segmentation
%% JOIN images ON segmentation.t = images.t
%% WHERE machine=112 and (algo like "alkofahi_nofinalize" or algo like "twangSBDE" or (algo like "mine%" and k=4 and binarization like "RegionalGaussianOtsu"))
%% GROUP BY algo, k, binarization;
   GC & 147 & 1.02 & 0 & 6.0 & 3.3 & 0.2 \\
 Ours &  56 & 0.83 & 0.4 & 0.8 & 1.3 & 1.8 \\
TWANG & 140 & 1.35 & 0.1 & 4.9 & 1.3 & 0 \\
% END TABULAR SELECT replace(replace(replace(algo, 'alkofahi_nofinalize', 'GC...
\bottomrule
\end{tabularx}
\end{table}

Table~\ref{tab:experimentsSBDE} lists results on the SBDE dataset. None of the tested algorithms exhibits perfect segmentation quality. The seed-based methods have very low rate of false positives, however fail to detect at least 5\% of the nuclei. The missing nuclei of GC are mainly at the far end of the visual axis, because GC has no mechanism to cope with the variations of illumination. The quality of our algorithm is similar or better than the other methods, only the number of split nuclei is elevated.

Our algorithm is again best regarding time. However, the running time varied between \SI{10}{\second} and \SI{150}{\second}. Running times of a segmentation algorithm can depend on several factors. 
The running time of our method is primarily affected by the number of clustered nuclei and the image resolution. The former is reflected in the number of partitioner calls, the latter increases the time needed per single partitioner call due to a larger graph. The running time variance of our method is caused by the variance in clustered objects for this data set, which ranges from 200 to 700 nuclei.
Errors in an image analysis pipeline are inevitable. An approach to cope with this fact is to quantify the uncertainty in the outcome of an operator \cite{stegmaier12challenges}. Our algorithm could be augmented by interpreting the score of our nuclei model as an uncertainty value. The error rate could then be reduced for narrow decisions of the nuclei model. A manual analysis of the errors in one image of the set showed that this was the case for 25\% of the errors made. 15\% of the errors could be attributed to the graph partitioner. The remaining errors would require a more exact nuclei model to be eliminated.

\section{Discussion and Future Work}
In this work we have presented a novel algorithm to split clustered nuclei by combining graph partitioning with a simple object model. We have shown the performance of our method on two sets of synthetic cell nuclei images. Our method is very fast and performs qualitatively comparable or better to other cell nuclei segmentation techniques.

Future work is necessary for a better automation. The required user input should be reduced, e.g. by unsupervised learning of the model parameters. This would allow the model to include more features, which could further improve the quality.
Another approach to reducing the user input, but also to allow more sophisticated binarization methods, would be to find a unifying model which can explain all used parameters. For an example, the parameters $\sigma_{\mathrm{grad}}$ and $\sigma_{\mathrm{bin}}$, used in the computation of the edge weights, should be deduced from the parameters of the automatically computed image model. We tried to find such a relationship, but it did not work in our experiments.

The anisotropic resolution was one of the biggest challenges for this work. The ratio between spacing in axial direction to the lateral directions was as high as $5$ in one of our datasets. This negatively affected several aspects of our method. For example, the approximation of sphericity is hard to obtain, as a nucleus extends only about $4$ slices in the axial direction. This can make it difficult to distinguish between a sphere, and a half-sphere. One solution for this issue would be to include more features in the nucleus model, for example the extent of the object in $z$-direction.

The anisotropic resolution is especially problematic for the gradient-based edge weighting method. Here, the edge weights are a measure for the probability of a change in intensity. However, such a change has a higher probability when moving along the $z$-axis, because pixels along this axis have a higher spacing. As a result, the weights for the respective edges are too small, favouring cuts of these edges. One idea to cope with this problem could be to adjust the parameter $\sigma_{\mathrm{grad}}$ accordingly. For example, one could use separate parameters depending on the edge orientation. 

This work has dealt with image segmentation of 3D images with anisotropic spacing. Another way to consider the images is as a stack of 2D images. This perception relates closely to the image formation process, as the image sensor which digitizes the real image is a 2D array of capacitors. One approach could be to segment all slices separately and then combine the detected 2D objects. This also has the advantage of an easy parallelization. Another idea is to separate the slices for the sphericity measure in the nucleus model: instead of using a three-dimensional sphericity, we could compute a two-dimensional roundness measure for each involved slice.

From a computer scientist's point of view, the problem in the development of image segmentation algorithms is that the problem is ill-posed \cite{danek12thesis, khairy11reconstructing}. For a well-posed problem, a solution exists, is unique, and is stable, i.e. it depends continuously on the data \cite{hadamard1902problemes}. The image segmentation problem does not have a unique solution, because the digitization process is not injective: multiple different images can be mapped to the same digital image. This is especially the case for the segmentation of live cell nuclei in 3D fluorescence microscopy. The low imaging quality and the anisotropic resolution lead to ambiguities. Even human experts can differ on their choice of the detection and/or the boundary of an object \cite{coelho09nuclear, danek12thesis}.

It is difficult to design a segmentation algorithm when the correct solution is not defined. This problem results from the lack of ground truth for large-scale image datasets. How should the quality of an algorithm be evaluated? How can the quality be optimized? How can it be compared to other algorithms? 
Some authors resort to subjective quality comparison \cite{stegmaier14fast}. Most either compare against hand-segmented datasets or synthetic benchmark datasets \cite{stegmaier16generating}. The latter choice is also the path taken in this work. 
%Another approach could be to evaluate downstream image analysis tasks. For example, we could measure the effect of different segmentation results on the tracking quality.
We have optimized our algorithm and the parameters with two sets of synthetic images. While our method shows superior results on these instances, its performance on real-world images has yet to be evaluated.
\newpage
\bibliographystyle{plainnat}
\bibliography{lit}
\end{document}